\documentclass{article}
\pdfoutput=1
\usepackage[table]{xcolor}

\usepackage{todonotes}
\usepackage{graphicx} %
\usepackage{multirow}
\usepackage[preprint]{corl_2024}
\usepackage{subcaption}
\usepackage[tableposition=top]{caption}

\usepackage{amsmath}
\usepackage{array}
\usepackage{authblk}
\usepackage[subtle]{savetrees}
\usepackage[all]{nowidow}

\usepackage{enumitem}
\usepackage{wrapfig}
\usepackage[all]{nowidow}
\usepackage{svg}
\usepackage{verbatim}
\usepackage{booktabs}

\makeatletter
\renewcommand\AB@affilsepx{, \protect\Affilfont}
\makeatother

\renewcommand*{\Affilfont}{\normalsize}
\newcommand{\algoName}{GeoMatch++}

\definecolor{lightblue}{rgb}{0.93,0.95,1.0}

\title{GeoMatch++: Morphology Conditioned Geometry Matching for Multi-Embodiment Grasping}
\vspace{-2cm}
\author[1]{\textbf{Yunze Wei}}
\author[1,2]{\textbf{Maria Attarian}}
\author[1]{\textbf{Igor Gilitschenski}}
\affil[1]{University of Toronto}
\affil[2]{Google DeepMind}

\date{September 2024}

\begin{document}

\maketitle

\vspace{-1cm}
\begin{abstract} Despite recent progress on multi-finger dexterous grasping, current methods focus on single grippers and unseen objects, and even the ones that explore cross-embodiment, often fail to generalize well to unseen end-effectors. This work addresses the problem of dexterous grasping generalization to unseen end-effectors via a unified policy that learns correlation between gripper morphology and object geometry. Robot morphology contains rich information representing how joints and links connect and move with respect to each other and thus, we leverage it through attention to learn better end-effector geometry features. Our experiments show an average of \textbf{9.64\%}  increase in grasp success rate across 3 out-of-domain end-effectors compared to previous methods. 
\end{abstract}


\footnotetext[0]{Correspondence emails: lulu.wei@mail.utoronto.ca, jmattarian@google.com}

\section{Introduction}

As we aspire to solve more dexterous tasks in robotics, multi-finger grasping becomes of increasing importance. However, the varying degrees of freedom (DoF) of end-effectors and high multimodality of grasping modes depending on both end-effectors and objects, still pose open challenges. Previous works in grasping focus on parallel grippers~\citep{Sundermeyer2021, Chisari2024, Fang2023}, a single multi-finger gripper~\citep{Weng2024a, Xu2023, Xu2024, Mayer2022}, or a shared policy for multiple dexterous grippers~\citep{attarian2023geometry,li2023gendexgrasp,Shao2020, Li2022}. However, even methods that explore cross-embodiment mostly focus on generalization to unseen objects, and still show limited zero-shot generalization to unseen grippers. 

In this work, we propose GeoMatch++, a multi-embodiment grasping method which improves out-of-domain generalization on unseen grippers by leveraging robot morphology. Intuitively, robot morphology is essential to grasping -- various end-effectors may have a different number of fingers, but fingertips and palm tend to be the most frequent contact regions. Thus, we hypothesize that learning good morphology embeddings can lead to a transferable grasping policy between different robots. %GeoMatch++ learns features and correspondence of point clouds and morphology using to autoregressively predict contact points for generating diverse stable grasps.
%To achieve this, we model gripper morphologies as graphs from their Unified Robot Description Format (URDF) representations, which contain the kinematic chain that describes the connectivity between joints and links, and their respective parameters. We learn embeddings for the point cloud and morphology graphs using Graph Convolutional Networks (GCN)~\citep{kipf2016semi}, then learn cross-correlation between them using transformer modules ~\citep{vaswani2017attention} with self-attention and cross-attention. Finally, we generate contact point predictions through autoregression.%
Our main contribution is learning geometry correlation features between objects and end-effector morphology, which improve out-of-domain grasp success by $9.64\%$ compared to previous methods, and our method showcases a minimal decrease in performance compared to in-domain evaluation.

\section{Related Work}
\label{sec:related_work}

\textbf{Dexterous Grasping:}
    Works focused on grasping for multi-finger grippers either train an end-to-end model to predict gripper pose directly \citep{Weng2024a, Xu2024, Mayer2022} or learn a contact map distribution before computing the final grasp \citep{attarian2023geometry, li2023gendexgrasp, Shao2020, Li2022}. Many of these methods are either constrained to one end-effector while others can generalize to unseen grippers, yet rarely incorporate gripper morphology explicitly, which better represents how complex multiple DoF grippers move during grasping. TAX-Pose~\citep{pan2023tax} is a recent method that learns a task-specific pose relationship between target objects to address manipulation tasks that involve multiple objects. The authors are inspired by Deep Closest Point (DCP)~\citep{Wang2019}, which proposes using transformers~\citep{Vaswani2017} to learn a matching between point clouds, and show that attention is also beneficial to the grasping problem. Instead of capturing the attention between point clouds, we propose using self-attention and cross-attention between object point cloud and end-effector morphology to learn a transferable grasping policy. Our work extends GeoMatch~\citep{attarian2023geometry}, which uses Graph Convolutional Networks (GCN)~\citep{Kipf2017} to learn object and robot geometries then performs autoregressive matching to predict object-robot contact points, via incorporating such morphology self and cross-attention.

\begin{figure}[t]
    \centering
    \begin{subfigure}{0.35\linewidth}
        \centering
        \includegraphics[width=\linewidth]{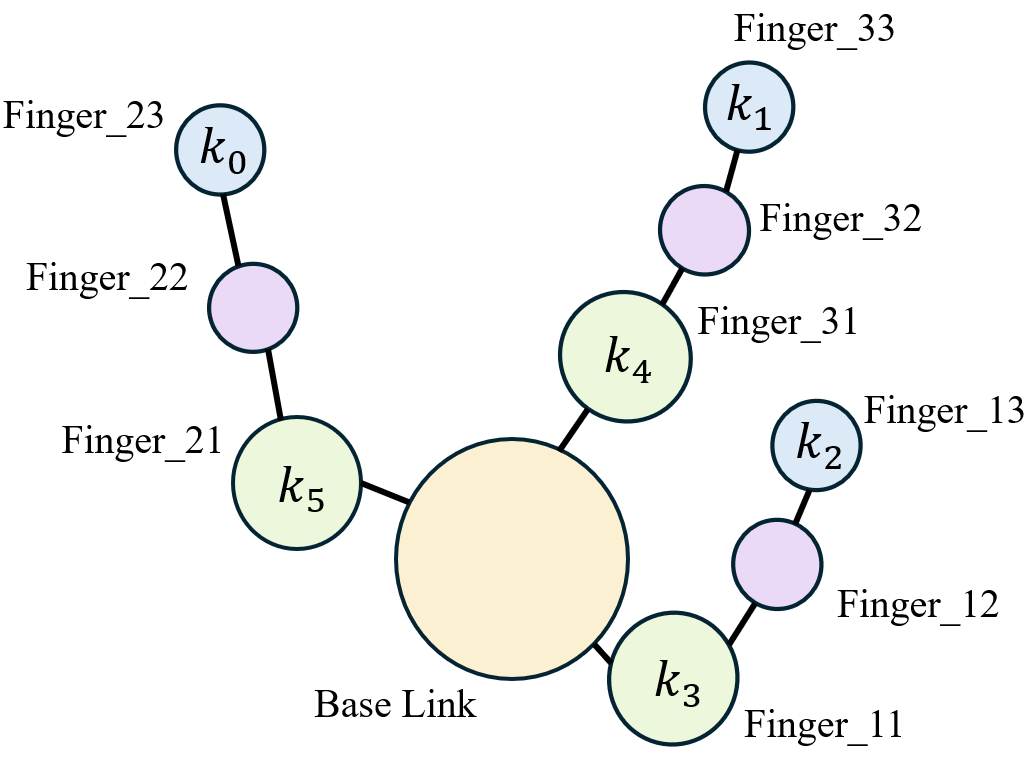}
    \end{subfigure}
    \hspace{1cm}
    \begin{subfigure}{0.3\linewidth}
        \centering
        \includegraphics[width=\linewidth]{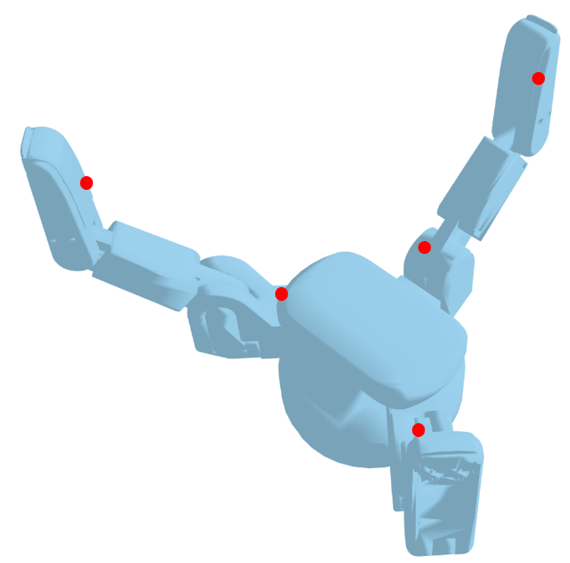}
    \end{subfigure}
    \caption{Sample morphology graph for Barrett hand with labelled keypoints.}
    \label{fig:combined morph}
    \vspace{-0.4cm}
\end{figure}

\textbf{Robot Morphology:}
    Robot morphology has been explored in other robotics control tasks in policy learning and imitation learning to generalize zero-shot to new tasks and agents~\citep{gupta2022metamorph, Wang2018, kurin2021my, Blake2021, Sferrazza2024}. A notable example is NerveNet~\citep{Wang2018} which explicitly models the structure of a modular agent as a graph, 
    %which stores the observation vectors for the body and joints of the agent,% 
    and propagates messages between nodes of the agent to train a reinforcement learning (RL) policy. 
    %that can generalize zero-shot to new tasks%. 
    Prior work has also explored robot structure as an inductive bias for transformers: MetaMorph~\citep{gupta2022metamorph} conditions a transformer on morphology and learns a universal controller while %for the UNIMAL design space of modular robots of 15-20 DoFs;% 
    Body Transformer (BoT)~\citep{Sferrazza2024} considers agent sensors and actuators as graph nodes and modifies attention masking to leverage morphology of the agent's structure. 
    %Evaluation in an imitation learning setting shows that their architecture outperforms MLP and vanilla transformers in online RL.%
    We show that morphology similarly leads to an improvement in generalization for cross-embodiment dexterous grasping.

\section{Method}
\label{sec:method}
\begin{figure}[ht]
    \centering
    \begin{subfigure}{0.7\linewidth}
        \centering
        \includegraphics[width=\linewidth]{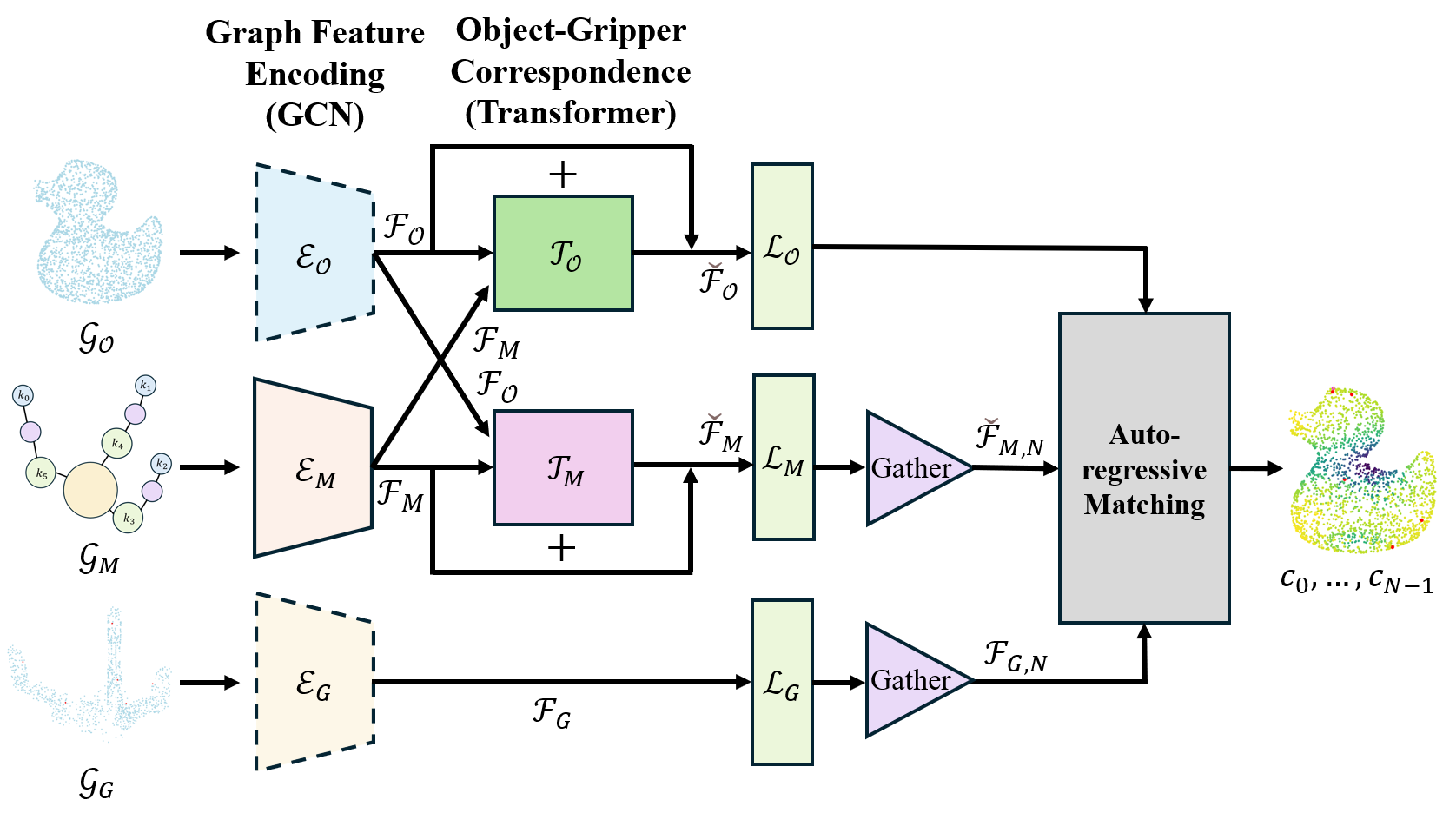}
        \caption{Model architecture}
        \label{fig:architecture}
    \end{subfigure}
    %\hfill
    \begin{subfigure}{0.29\linewidth}
        \centering
        \includegraphics[width=\linewidth]{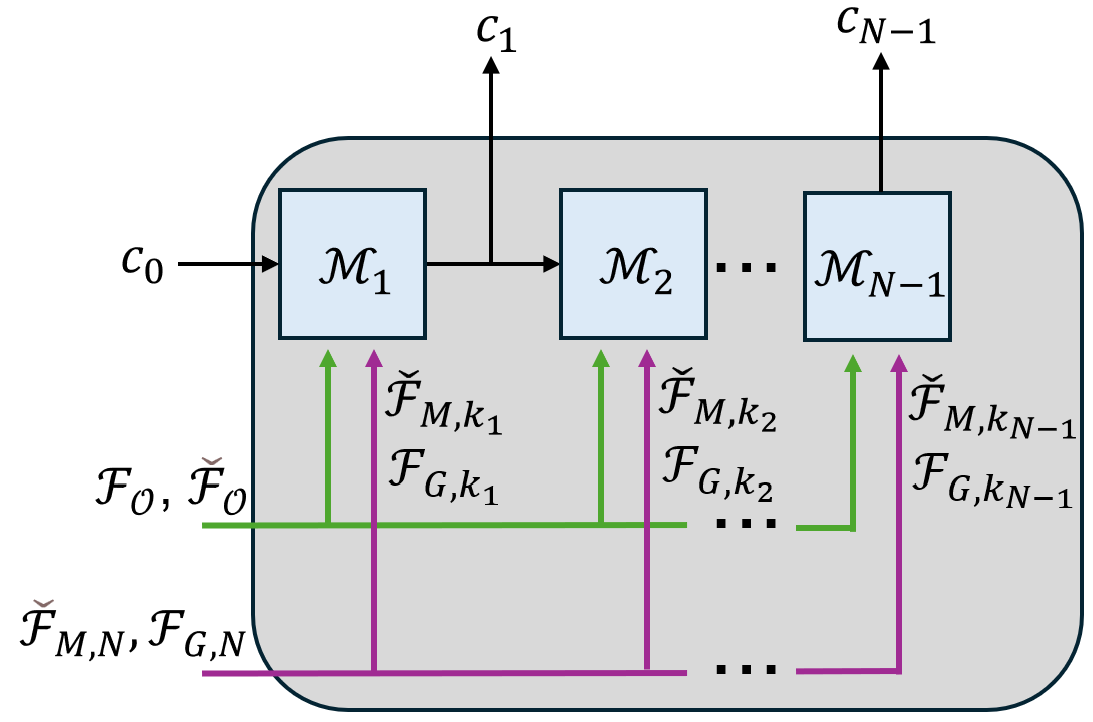}
        \caption{Autoregressive module}
        \label{fig:autoregressive}
    \end{subfigure}
    \caption{Architecture for GeoMatch++. GCNs learn latent features for object and gripper point clouds, and end-effector morphology. Features are passed into transformer modules to learn the object-gripper correspondence. Autoregressive matching predicts final contact points using MLP layers.}
    \label{fig:combined}
\end{figure}

Our model (Fig. \ref{fig:combined}) learns a multi-embodiment policy that generates diverse grasps for dexterous grippers for both unseen objects and end-effectors. Operating under the same problem formulation as \citep{attarian2023geometry}, we match $N=6$ pre-defined keypoints on the end-effector $k_0, \ldots, k_{N-1}$ to predicted contact points on the object $c_0, \ldots, c_{N-1}$. %, such that a stable grasp is achieved when the two sets are in contact. 
Our model encodes graph features for the object point cloud $\mathcal{G}_O$, gripper point cloud $\mathcal{G}_G$, and the graph representing the morphology of the gripper $\mathcal{G}_M$ using GCNs. Transformer modules perform self-attention and cross-attention to capture global correspondence between the object and end-effector. Finally, the model autoregressively predicts contact points using the latent embeddings. %We use $N=6$ for the number of keypoint-contact pairs. The keypoints are chosen to lie on different links to capture diverse morphological information and to be semantically consistent across end-effectors, but otherwise satisfy no other constraint.%
\vspace{-0.1cm}
\subsection{Dataset}
\label{subsec:dataset}
We use a subset of the MultiDex dataset synthesized by \cite{li2023gendexgrasp} using force closure optimization \cite{liu2021synthesizing}. The dataset contains 5 high-DoF multi-finger grippers, EZGripper,
Barrett, Robotiq-3F, Allegro, and ShadowHand, and 58 household objects from the ContactDB \cite{brahmbhatt2019contactdb} and YCB \cite{Calli2017} datasets. We train on 50,802 grasps, represented by poses consisting of translation, rotation, and joint angles of the gripper.

\vspace{-0.1cm}
\subsection{Graph Representation}
\label{subsec:graphrep}
\textbf{Object and End-effector Point Clouds:} Object and end-effector point clouds are represented as graphs $ \mathcal{G}_O = (\mathcal{V}_O, \mathcal{E}_O) $, and $\mathcal{G}_G = (\mathcal{V}_G, \mathcal{E}_G)$. Each point is represented as its 3D coordinates $\mathbf{p_{i}} = (x_i, y_i, z_i)\in\mathcal{R}^3$. The graph is constructed by sampling $S_O=2048$ points for the object mesh and $S_G=1000$ points from the end-effector mesh. Prior to sampling, the end-effector is set to a canonical rest pose that has zero root translation, zero root rotation, and all joints set to the middle of their joint limits.

\textbf{End-effector Morphology Representation:} 
The end-effector's kinematic chain, which contains information about link-joint connections and parameters, is obtained from the Universal Robot Description Format (URDF) and constructed as a graph $\mathcal{G}_M = (\mathcal{V}_M, \mathcal{E}_M)$.
In our setup, nodes $\mathcal{V}_M$ are links and edges $\mathcal{E}_M$ are joints (Fig.~\ref{fig:combined morph}). 
The graph features consist of offset, link centre of mass, and link size. 
Offset represents the translation between the coordinate frames of two connected links. 
Link centre of mass is estimated via computing the least volume rectangular bounding boxes around the link mesh and finding its mean coordinate on each axis.
Finally, link size is the length, width, and height of the bounding box. The coordinate frame of the centres of mass and the scale of the link sizes are all geometrically consistent with object and end-effector point clouds.
Only the offset is encoded relative to two connected nodes. Due to varied DoFs of end-effectors, $\mathcal{G}_M$ is zero-padded to $S_M = 32$ to enable batch processing. More details are given in Appendix \ref{appendix:morph graph}.
%Therefore, morphology features for all end-effectors can be represented as $\mathcal{P_M}\in\mathcal{R}^{9\times{S_M}}$.

\vspace{-0.2cm}

\subsection{Architecture}

\textbf{Graph Feature Encoding:}
The model uses three separate GCNs to generate latent embeddings of dimension $n=512$ for $\mathcal{G}_O$, $\mathcal{G}_G$, and $\mathcal{G}_M$. We use $\mathcal{F}_O$, $\mathcal{F}_G$, and $\mathcal{F}_M$ to represent the latent embeddings. We use pretrained weights from GeoMatch~\cite{attarian2023geometry} for $\mathcal{F}_O$ and $\mathcal{F}_G$ and freeze them during training, as empirically this shows the best performance.
$\mathcal{G}_M$ is novel to our model and trained from scratch. $\mathcal{G}_M$ is zero-padded to account for different DoFs in the end-effectors which does not pose an issue given GCN's property of only aggregating features of a node's direct neighbourhood.

\textbf{Object-Gripper Correspondence:} 
We use two transformer modules with self-attention and cross-attention to learn correspondence between the latent embeddings for object features $\mathcal{F}_O$ and morphology $\mathcal{F}_M$. Following~\citet{Wang2019}, we consider the output of the transformer as a residual term and add it to the GCN encoding:
\begin{equation}
\begin{aligned}
\hat{\mathcal{F}}_O = \mathcal{F}_O + \mathcal{T_O}(\mathcal{F}_O, \mathcal{F}_M) \hspace{0.5em} \in \mathcal{R}^{n \times S_O} 
\quad &
\hat{\mathcal{F}}_M = \mathcal{F}_M + \mathcal{T_M}(\mathcal{F}_M, \mathcal{F}_O) \hspace{0.5em} \in \mathcal{R}^{n \times S_M}
\end{aligned}
\end{equation}
This operation modifies features $\mathcal{F}_O$ and $\mathcal{F}_M$ such that they are aware of the correlation between object and morphology. Then, linear layers downsample the embeddings for further processing.

\textbf{Autoregressive Matching:}
We modify the autoregressive module from~\cite{attarian2023geometry} to incorporate morphology encodings. 
We gather $\mathcal{F}_G$ and $\hat{\mathcal{F}}_M$ to obtain only the embedding corresponding to the $N$ keypoints. Each layer $\mathcal{M}_i$ in autoregressive matching is an MLP that predicts contact point $c_i$ from the concatenation of the full object embedding $\hat{\mathcal{F}}_O$, gathered embeddings $\mathcal{F}_{G, N}$ and $\hat{\mathcal{F}}_{M, N}$ repeated $S_O = 2048$ times, and the contact points $c_0, \ldots, c_{i-1}$ from the previous layers. $c_0$ is predicted from the unnormalized likelihood contact maps, further explained in Section \ref{subsec:losses}. Although only the $i$-th feature of $\hat{\mathcal{F}}_{M, N}$ is used in layer $M_i$, $\hat{\mathcal{F}}_O$ contains information about the entire end-effector morphology through cross-attention.

\subsubsection{Losses}
\label{subsec:losses}
We use the same loss functions as \citet{attarian2023geometry}, consisting of the Geometric Embedding Loss and Predicted Contact Loss, with modifications described below. For more details, we refer the reader to the paper.

\textbf{Geometric Embedding Loss:} We calculate the BCE loss between the predicted unnormalized likelihood contact maps for each pair of object vertex $v_o$ and keypoint $k_i$, and the ground truth contact maps $C_O(v_o, k_i)$. 
Instead of learning the contact maps using GCN encodings as done in GeoMatch~\citep{attarian2023geometry}, we use the dot product between the object-gripper correspondence transformer output of the object point cloud and the GCN embeddings of the gripper point cloud.

\textbf{Predicted Contact Loss:} We use the same predicted contact loss as~\citep{attarian2023geometry} to train autoregressive matching contact point predictions.

\section{Experiments and Discussion}
\label{sec:experiments}

We use the same evaluation setup as~\citep{attarian2023geometry, li2023gendexgrasp} which leverages IsaacGym to measure grasp success rate and diversity. Grasp success rate is calculated over four grasps per object-gripper pairs, and diversity is measured as the standard deviation of the joint angles of successful grasps.

\vspace{-0.2cm}
\subsection{Out-of-domain evaluation}

The model is evaluated on out-of-domain grippers by training on 4 out of 5 grippers, and testing using the unseen gripper with 10 unseen objects. We choose to compare results with two recent methods that focus on multi-gripper dexterous grasping, GeoMatch \cite{attarian2023geometry} and GenDexGrasp \cite{li2023gendexgrasp}.

\rowcolors{1}{}{lightblue}
\begin{table}[t]

\centering
\resizebox{0.9\columnwidth}{!} {%
\begin{tabular}{ c || c | c | c | c | c | c | c }
\hline
  \multirow{2}{*}{\textbf{Method}} &
  \multicolumn{4}{c|}{\textbf{Success (\%) $\uparrow$}} & \multicolumn{3}{c}{\textbf{Diversity (rad) $\uparrow$ }}\\ \cline{2-8}
 & ezgripper & barrett & shadowhand & \textbf{Mean} & ezgripper & barrett & shadowhand \\  
 \midrule
  GeoMatch \cite{attarian2023geometry} & 55.0 & 60.0 & 67.5 & 60.83 & 0.185 & 0.259 & \textbf{0.235}  \\
  GenDexGrasp \cite{li2023gendexgrasp} & 38.59 & 70.31 & \textbf{77.19} & 62.03 & \textbf{0.248} &  0.267 & 0.207  \\
 \textbf{\algoName} (ours) & \textbf{67.5} & \textbf{77.5} & 70.0 & \textbf{71.67} & 0.208 & \textbf{0.378} & 0.184 \\
 \hline
\end{tabular}
}
\caption{Out-of-domain success rate and diversity comparisons with GeoMatch and GenDexGrasp}
\label{tab:out of domain}
\vspace{-0.4cm}
\end{table}

\begin{figure}[h!]
    \centering
    \includegraphics[width=0.9\linewidth]{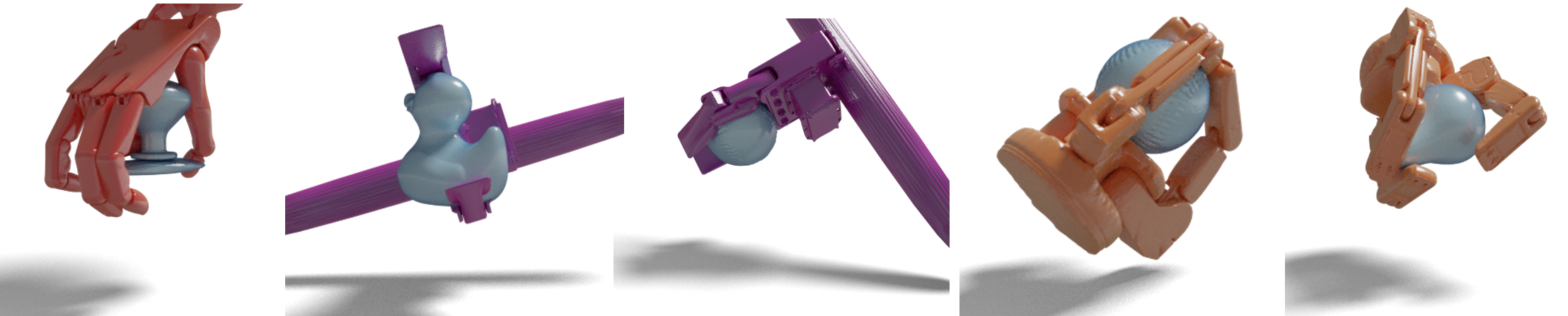}
    \caption{Qualitative grasp results on unseen grippers.} 
    \label{fig:results}
\end{figure}

Our model shows significant improvement in out-of-domain generalization, having a mean success rate of \textbf{71.67\%} and a mean grasp diversity of $0.257$. GeoMatch++ outperforms the mean success rate of GeoMatch~\citep{attarian2023geometry} by \textbf{10.84\%} and GenDexGrasp \citep{li2023gendexgrasp} by \textbf{9.64\%} (Table \ref{tab:out of domain}). Furthermore, our mean out-of-domain performance is only \textbf{3.33\%} lower than in-domain ($75.0\%$, Appendix \ref{sec:in-domain eval}), demonstrating the method's strength in generalizing to new grippers. Sample grasps are rendered in Figure \ref{fig:results}.

\vspace{-0.2cm}

\subsection{Ablations}
\label{subsec:ablations}

\textbf{Q1: What is the importance of starting training from good point cloud embeddings?}
We train ablations where weights of $ \mathcal{G}_O$ and $\mathcal{G}_G$ are trained from scratch, pretrained and fintuned, or pretrained and frozen. Empirically, freezing the pretrained weights achieves the best success rate. In particular, we note that training from scratch suffers a large drop in success rate ($24.97\%$ $\downarrow$) (Appendix \ref{appendix:weights}). 

\textbf{Q2: Does including robot morphology improve out-of-domain generalization?} To examine the role of end-effector morphology in generalization, we remove morphology completely and add transformer modules between the object and robot point clouds instead. 
We find that mean success rate of our final method (including morphology) is $22.51\%$ higher than without morphology (Appendix \ref{appendix:morphology or not}).

\textbf{Q3: What is the contribution of different morphology features?} The relative importance of features of the robot morphology graph is examined through using different combinations of morphological features in $\mathcal{G}_M$. We run ablations for joints only features (relative offset, joint axis, joint limits) and links only features (absolute origin coordinates, centre of mass, size of bounding box). Our final selection of features, with a combination of relative offset and link coordinate information, achieves the best results (Appendix \ref{appendix:morphology features}).

\section{Conclusion}
In this paper we propose a novel method, GeoMatch++, that leverages robot morphology to improve out-of-domain generalization to unseen grippers. We demonstrate that learning robot link and joint features and the object-morphology correlation are important for achieving high grasp success rates out-of-domain, outperforming baseline by $9.64\%$. 
%We show that our method outperforms baseline by $9.64\%$ out-of-domain, and that success rate only drops $3.33\%$ out-of-domain compared to in-domain. 
We hope this work is a step forward towards zero-shot generalization to unseen grippers in real robot settings.

\bibliography{geomatch++}

\begin{thebibliography}{24}
\providecommand{\natexlab}[1]{#1}
\providecommand{\url}[1]{\texttt{#1}}
\expandafter\ifx\csname urlstyle\endcsname\relax
  \providecommand{\doi}[1]{doi: #1}\else
  \providecommand{\doi}{doi: \begingroup \urlstyle{rm}\Url}\fi

\bibitem[Sundermeyer et~al.(2021)Sundermeyer, Mousavian, Triebel, and
  Fox]{Sundermeyer2021}
M.~Sundermeyer, A.~Mousavian, R.~Triebel, and D.~Fox.
\newblock Contact-graspnet: Efficient 6-dof grasp generation in cluttered
  scenes.
\newblock In \emph{2021 IEEE International Conference on Robotics and
  Automation (ICRA)}, pages 13438--13444. IEEE, 2021.

\bibitem[Chisari et~al.(2024)Chisari, Heppert, Welschehold, Burgard, and
  Valada]{Chisari2024}
E.~Chisari, N.~Heppert, T.~Welschehold, W.~Burgard, and A.~Valada.
\newblock Centergrasp: Object-aware implicit representation learning for
  simultaneous shape reconstruction and 6-dof grasp estimation.
\newblock \emph{IEEE Robotics and Automation Letters}, 2024.

\bibitem[Fang et~al.(2023)Fang, Wang, Fang, Gou, Liu, Yan, Liu, Xie, and
  Lu]{Fang2023}
H.-S. Fang, C.~Wang, H.~Fang, M.~Gou, J.~Liu, H.~Yan, W.~Liu, Y.~Xie, and
  C.~Lu.
\newblock Anygrasp: Robust and efficient grasp perception in spatial and
  temporal domains.
\newblock \emph{IEEE Transactions on Robotics}, 2023.

\bibitem[Weng et~al.(2024)Weng, Lu, Kragic, and Lundell]{Weng2024a}
Z.~Weng, H.~Lu, D.~Kragic, and J.~Lundell.
\newblock Dexdiffuser: Generating dexterous grasps with diffusion models, 2024.
\newblock URL \url{https://arxiv.org/abs/2402.02989}.

\bibitem[Xu et~al.(2023)Xu, Wan, Zhang, Liu, Shan, Shen, Wang, Geng, Weng,
  Chen, et~al.]{Xu2023}
Y.~Xu, W.~Wan, J.~Zhang, H.~Liu, Z.~Shan, H.~Shen, R.~Wang, H.~Geng, Y.~Weng,
  J.~Chen, et~al.
\newblock Unidexgrasp: Universal robotic dexterous grasping via learning
  diverse proposal generation and goal-conditioned policy.
\newblock In \emph{Proceedings of the IEEE/CVF Conference on Computer Vision
  and Pattern Recognition}, pages 4737--4746, 2023.

\bibitem[Xu et~al.(2024)Xu, Wei, Zheng, Wu, and Zheng]{Xu2024}
G.-H. Xu, Y.-L. Wei, D.~Zheng, X.-M. Wu, and W.-S. Zheng.
\newblock Dexterous grasp transformer.
\newblock In \emph{Proceedings of the IEEE/CVF Conference on Computer Vision
  and Pattern Recognition (CVPR)}, pages 17933--17942, June 2024.

\bibitem[Mayer et~al.(2022)Mayer, Feng, Deng, Shi, Chen, and Knoll]{Mayer2022}
V.~Mayer, Q.~Feng, J.~Deng, Y.~Shi, Z.~Chen, and A.~Knoll.
\newblock Ffhnet: Generating multi-fingered robotic grasps for unknown objects
  in real-time.
\newblock In \emph{2022 International Conference on Robotics and Automation
  (ICRA)}, pages 762--769, 2022.

\bibitem[Attarian et~al.(2023)Attarian, Asif, Liu, Hari, Garg, Gilitschenski,
  and Tompson]{attarian2023geometry}
M.~Attarian, M.~A. Asif, J.~Liu, R.~Hari, A.~Garg, I.~Gilitschenski, and
  J.~Tompson.
\newblock Geometry matching for multi-embodiment grasping.
\newblock In \emph{Conference on Robot Learning}, pages 1242--1256. PMLR, 2023.

\bibitem[Li et~al.(2023)Li, Liu, Li, Geng, Zhu, Yang, and
  Huang]{li2023gendexgrasp}
P.~Li, T.~Liu, Y.~Li, Y.~Geng, Y.~Zhu, Y.~Yang, and S.~Huang.
\newblock Gendexgrasp: Generalizable dexterous grasping.
\newblock In \emph{2023 IEEE International Conference on Robotics and
  Automation (ICRA)}, pages 8068--8074. IEEE, 2023.

\bibitem[Shao et~al.(2020)Shao, Ferreira, Jorda, Nambiar, Luo, Solowjow, Ojea,
  Khatib, and Bohg]{Shao2020}
L.~Shao, F.~Ferreira, M.~Jorda, V.~Nambiar, J.~Luo, E.~Solowjow, J.~A. Ojea,
  O.~Khatib, and J.~Bohg.
\newblock Unigrasp: Learning a unified model to grasp with multifingered
  robotic hands.
\newblock \emph{IEEE Robotics and Automation Letters}, 5\penalty0 (2):\penalty0
  2286--2293, 2020.

\bibitem[Li et~al.(2022)Li, Baron, Zhang, and Rojas]{Li2022}
K.~Li, N.~Baron, X.~Zhang, and N.~Rojas.
\newblock Efficientgrasp: A unified data-efficient learning to grasp method for
  multi-fingered robot hands.
\newblock \emph{IEEE Robotics and Automation Letters}, 7\penalty0 (4):\penalty0
  8619--8626, 2022.

\bibitem[Pan et~al.(2023)Pan, Okorn, Zhang, Eisner, and Held]{pan2023tax}
C.~Pan, B.~Okorn, H.~Zhang, B.~Eisner, and D.~Held.
\newblock Tax-pose: Task-specific cross-pose estimation for robot manipulation.
\newblock In \emph{Conference on Robot Learning}, pages 1783--1792. PMLR, 2023.

\bibitem[Wang and Solomon(2019)]{Wang2019}
Y.~Wang and J.~M. Solomon.
\newblock Deep closest point: Learning representations for point cloud
  registration.
\newblock In \emph{Proceedings of the IEEE/CVF International Conference on
  Computer Vision (ICCV)}, October 2019.

\bibitem[Vaswani(2017)]{Vaswani2017}
A.~Vaswani.
\newblock Attention is all you need.
\newblock \emph{Advances in Neural Information Processing Systems}, 2017.

\bibitem[Kipf and Welling(2017)]{Kipf2017}
T.~N. Kipf and M.~Welling.
\newblock Semi-supervised classification with graph convolutional networks.
\newblock In \emph{International Conference on Learning Representations
  (ICLR)}, 2017.

\bibitem[Gupta et~al.(2022)Gupta, Fan, Ganguli, and
  Fei-Fei]{gupta2022metamorph}
A.~Gupta, L.~Fan, S.~Ganguli, and L.~Fei-Fei.
\newblock Metamorph: Learning universal controllers with transformers.
\newblock In \emph{International Conference on Learning Representations}, 2022.

\bibitem[Wang et~al.(2018)Wang, Liao, Ba, and Fidler]{Wang2018}
T.~Wang, R.~Liao, J.~Ba, and S.~Fidler.
\newblock Nervenet: Learning structured policy with graph neural networks.
\newblock In \emph{International conference on learning representations}, 2018.

\bibitem[Kurin et~al.(2021)Kurin, Igl, Rockt{\"a}schel, Boehmer, and
  Whiteson]{kurin2021my}
V.~Kurin, M.~Igl, T.~Rockt{\"a}schel, W.~Boehmer, and S.~Whiteson.
\newblock My body is a cage: the role of morphology in graph-based incompatible
  control.
\newblock In \emph{International Conference on Learning Representations}, 2021.

\bibitem[Blake et~al.(2021)Blake, Kurin, Igl, and Whiteson]{Blake2021}
C.~Blake, V.~Kurin, M.~Igl, and S.~Whiteson.
\newblock Snowflake: Scaling gnns to high-dimensional continuous control via
  parameter freezing.
\newblock \emph{Advances in Neural Information Processing Systems},
  34:\penalty0 23983--23992, 2021.

\bibitem[Sferrazza et~al.(2024)Sferrazza, Huang, Liu, Lee, and
  Abbeel]{Sferrazza2024}
C.~Sferrazza, D.-M. Huang, F.~Liu, J.~Lee, and P.~Abbeel.
\newblock Body transformer: Leveraging robot embodiment for policy learning.
\newblock \emph{arXiv preprint arXiv:2408.06316}, 2024.

\bibitem[Liu et~al.(2021)Liu, Liu, Jiao, Zhu, and Zhu]{liu2021synthesizing}
T.~Liu, Z.~Liu, Z.~Jiao, Y.~Zhu, and S.-C. Zhu.
\newblock Synthesizing diverse and physically stable grasps with arbitrary hand
  structures using differentiable force closure estimator.
\newblock \emph{IEEE Robotics and Automation Letters}, 7\penalty0 (1):\penalty0
  470--477, 2021.

\bibitem[Brahmbhatt et~al.(2019)Brahmbhatt, Ham, Kemp, and
  Hays]{brahmbhatt2019contactdb}
S.~Brahmbhatt, C.~Ham, C.~C. Kemp, and J.~Hays.
\newblock Contactdb: Analyzing and predicting grasp contact via thermal
  imaging.
\newblock In \emph{Proceedings of the IEEE/CVF conference on computer vision
  and pattern recognition}, pages 8709--8719, 2019.

\bibitem[Calli et~al.(2017)Calli, Singh, Bruce, Walsman, Konolige, Srinivasa,
  Abbeel, and Dollar]{Calli2017}
B.~Calli, A.~Singh, J.~Bruce, A.~Walsman, K.~Konolige, S.~Srinivasa, P.~Abbeel,
  and A.~M. Dollar.
\newblock Yale-cmu-berkeley dataset for robotic manipulation research.
\newblock \emph{The International Journal of Robotics Research}, 36\penalty0
  (3):\penalty0 261--268, 2017.

\bibitem[{Dawson-Haggerty et al.}()]{trimesh}
{Dawson-Haggerty et al.}
\newblock trimesh.
\newblock URL \url{https://trimesh.org/}.

\end{thebibliography}

\newpage

\appendix

\section{In-domain Evaluation}
\label{sec:in-domain eval}
In domain, our model's mean success rate across the 3 evaluated grippers is \textbf{75.0\%}, outperforming GenDexGrasp by \textbf{10.89\%} and being worse than GeoMatch by \textbf{4.19\%} (Table \ref{tab:in domain}). Despite the minor drop in performance compared to baseline, our model shows significant improvement in out-of-domain performance.

\rowcolors{1}{}{lightblue}
\begin{table}[h!]
\centering
\resizebox{0.9\columnwidth}{!} {%
\begin{tabular}{ c || c | c | c | c | c | c | c }
\hline
  \multirow{2}{*}{\textbf{Method}} &
  \multicolumn{4}{c|}{\textbf{Success (\%) $\uparrow$}} & \multicolumn{3}{c}{\textbf{Diversity (rad) $\uparrow$ }}\\ \cline{2-8}
 & ezgripper & barrett & shadowhand & \textbf{Mean} & ezgripper & barrett & shadowhand \\  
 \midrule
  GeoMatch \cite{attarian2023geometry} & 75.0 & \textbf{90.0} & 72.5 & \textbf{79.17} & 0.188 & 0.249 & 0.205 \\
  GenDexGrasp \cite{li2023gendexgrasp} & 43.44 & 71.72 & \textbf{77.03} & 64.11 & \textbf{0.238} & 0.248 & \textbf{0.211} \\
 \textbf{\algoName} (ours)  & \textbf{82.5} & 72.5 & 70.0 & 75.0 & 0.175 & \textbf{0.342} & 0.206 \\
 \hline
\end{tabular}
}
\caption{In-domain success rate and diversity comparisons with GeoMatch and GenDexGrasp}
\label{tab:in domain}
\end{table}

\section{Ablations}

\label{sec:Ablations}

We include results from ablation studies used to support the discussion in Section \ref{subsec:ablations}.

\subsection{What is the importance of starting training from good point cloud embeddings?}
\label{appendix:weights}

\global\rownum=0\relax
\rowcolors{1}{}{lightblue}
\begin{table}[h!]
\centering
\resizebox{0.9\columnwidth}{!} {%
\begin{tabular}{ c || c | c | c | c | c | c | c }
\hline
  \multirow{2}{*}{\textbf{Method}} &
  \multicolumn{4}{c|}{\textbf{Success (\%) $\uparrow$}} & \multicolumn{3}{c}{\textbf{Diversity (rad) $\uparrow$ }}\\ \cline{2-8}
 & ezgripper & barrett & shadowhand & \textbf{Mean} & ezgripper & barrett & shadowhand \\  
 \midrule
  From Scratch & 57.5 & 62.5 & 20.0 & 46.7 & 0.222 & 0.197 & 0.116 \\
Pretrained (Finetune) & \textbf{72.5} & \textbf{77.5} & 55.0 & 68.33 & \textbf{0.255} & 0.318 & \textbf{0.221} \\
\textbf{Pretrained (Freeze)} & 67.5 & \textbf{77.5} & \textbf{70.0} & \textbf{71.67} & 0.208 & \textbf{0.378} & 0.184 \\
 \hline
\end{tabular}
}
\caption{Comparison of weights for point cloud GCN embeddings}
\label{table:ablation - embedding}
\end{table}

\subsection{Does including robot morphology improve out-of-domain generalization?}
\label{appendix:morphology or not}

\global\rownum=0\relax
\rowcolors{1}{}{lightblue}
\begin{table}[h!]
\centering
\resizebox{0.9\columnwidth}{!} {%
\begin{tabular}{ c || c | c | c | c | c | c | c }
\hline
  \multirow{2}{*}{\textbf{Method}} &
  \multicolumn{4}{c|}{\textbf{Success (\%) $\uparrow$}} & \multicolumn{3}{c}{\textbf{Diversity (rad) $\uparrow$ }}\\ \cline{2-8}
 & ezgripper & barrett & shadowhand & \textbf{Mean} & ezgripper & barrett & shadowhand \\  
 \midrule
  Point Cloud Only & 27.5 & 70.0 & 50.0 & 49.16 & \textbf{0.270} & \textbf{0.429} & 0.141 \\
\textbf{PC and Morphology} & \textbf{67.5} & \textbf{77.5} & \textbf{70.0} & \textbf{71.67} & 0.208 & 0.378 & \textbf{0.184} \\
 \hline
\end{tabular}
}
\caption{Comparison of using only point clouds vs. using point clouds and morphology}
\label{table:ablation - PC vs morphology}
\end{table}

\subsection{What is the contribution of different morphology features?}
\label{appendix:morphology features}

\global\rownum=0\relax
\rowcolors{1}{}{lightblue}
\begin{table}[h!]
\centering
\resizebox{0.9\columnwidth}{!} {%
\begin{tabular}{ c || c | c | c | c | c | c | c }
\hline
  \multirow{2}{*}{\textbf{Method}} &
  \multicolumn{4}{c|}{\textbf{Success (\%) $\uparrow$}} & \multicolumn{3}{c}{\textbf{Diversity (rad) $\uparrow$ }}\\ \cline{2-8}
 & ezgripper & barrett & shadowhand & \textbf{Mean} & ezgripper & barrett & shadowhand \\  
 \midrule
  Joints Only & 62.5 & 72.5 & 62.5 & 65.83 & 0.209 & \textbf{0.390} & 0.199 \\
Links Only & 57.5 & 67.5 & 57.5 & 60.83 & \textbf{0.244} & 0.271 &  \textbf{0.215} \\
\textbf{Final} & \textbf{67.5} & \textbf{77.5} & \textbf{70.0} & \textbf{71.67} & 0.208 & 0.378 & 0.184\\
 \hline
\end{tabular}
}
\caption{Comparison of different morphology features}
\label{table:ablation - morphology feats}
\end{table}

\section{Morphology Graph Representation}
\label{appendix:morph graph}
We formulate the morphology graph from the URDF description of each end-effector. Nodes of the graph are links and edges are joints. We consider both revolute and fixed joints as edges. Two nodes are connected if they are respectively the parent and child link of a joint. Self-connections are added in the graph. The offset feature is obtained from the $<joint><origin><xyz>$ element of a joint. The feature is attributed to the child link of the joint. End-effectors may have a root link that is connected to a joint with multiple children links; in this case, the offset feature is attributed to the child link first listed in the kinematic chain. The least volume rectangular bounding boxes of links are estimated from the link meshes using the Trimesh library~\citep{trimesh}. We determine the morphology features most useful for learning empirically.

\section{Implementation Details}
\label{appendix:implementation}

We use $N=6$ for the number of keypoint-contact pairs. The keypoints are chosen to lie on different links to capture diverse morphological information and to be semantically consistent across end-effectors, but otherwise satisfy no other constraint. 

Experiments are conducted on a RTX3090 GPU. The model is trained using Adam with a learning rate of 0.00005 and betas of (0.9, 0.99), for 150 epochs with batch size 32. The parameters for the GCNs and autoregressive module are similar to GeoMatch~\citep{attarian2023geometry}. GCNs have 3 hidden graph convolution layers of dimension 256, and a final output linear layer of dimension 512. Each autoregressive MLP contains 3 hidden layers of dimension 256 and outputs a contact likelihood map of size 2048. We use the same parameters for the object-gripper correspondence transformers as the transformer module in DCP~\citep{Wang2019}, but with input dimensions of object point cloud size $S_O=2048$ and morphology graph size $S_M=32$.

We use the same inverse kinematics optimization and IsaacGym evaluation setup as~\citep{attarian2023geometry}.

\end{document}